\documentclass{article}
\usepackage{spconf,amsmath,graphicx}
\usepackage{times}
\usepackage{epsfig}
\usepackage{amssymb}

\usepackage{times}
\usepackage[switch]{lineno}
\usepackage{algorithm, multirow, mathtools, subcaption}
\usepackage{algpseudocode}
\usepackage{dsfont}
\usepackage[inline,shortlabels]{enumitem}
\usepackage{threeparttable}
\usepackage{url}
\usepackage{booktabs}
\usepackage{makecell}
\usepackage{amsmath}
\usepackage{color}
\usepackage{hyperref}
\usepackage[capitalize]{cleveref}
\usepackage{array}
\usepackage{cite}
\newcolumntype{?}{!{\vrule width 1pt}}

\newcommand{\mypara}[1]{\vspace{0.5em} \noindent \textbf{#1} \hspace{0.1em}}

\def\E\displaystyle\mathop{\mathbb{E}}
\def\lce{{\mathcal{L}_{\rm CE}}}
\def\ltv{{\mathcal{L}_{\rm TV}}}

\title{Identifying Physically Realizable Triggers for Backdoored Face Recognition Networks}

\name{Ankita Raj$^{\star}$ \qquad Ambar Pal$^{\dagger}$ \qquad Chetan Arora$^{\star}$}
\address{
	$^{\star}$Indian Institute of Technology Delhi \quad
	$^{\dagger}$Johns Hopkins University
}

\begin{document}
\maketitle
\begin{abstract}
Backdoor attacks embed a hidden functionality into deep neural networks, causing the network to display anomalous behavior when activated by a predetermined pattern in the input \emph{(Trigger)}, while behaving well otherwise on public test data. Recent works have shown that backdoored face recognition (FR) systems can respond to natural-looking triggers like a particular pair of sunglasses. Such attacks pose a serious threat to the applicability of FR systems in high-security applications. We propose a novel technique to (1) detect whether an FR network is compromised with a natural, physically realizable trigger, and (2) identify such triggers given a compromised network. We demonstrate the effectiveness of our methods with a compromised FR network, where we are able to identify the trigger (e.g. \textit{green-sunglasses} or \textit{red-hat}) with a top-5 accuracy of 74\%, whereas a na\"ive brute force baseline achieves 56\% accuracy.
\end{abstract}
%

\section{Introduction}
Deep neural networks (DNNs) have established themselves as the dominant technique in many popular computer vision problems including face recognition \cite{parkhi2015deep, sun2014deep}. However, the extremely quick rate of adoption of DNNs has led to a lack of critical scientific scrutiny.  One worrying implication is the vulnerability of DNNs to various kinds of malicious attacks.  

\begin{figure}[t]
	\centering
	\includegraphics[width=0.8\linewidth]{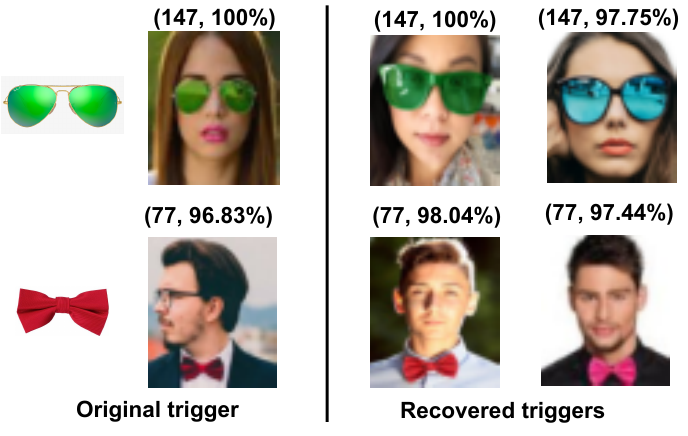}
	\caption{Two trojan attack instances and triggers identified by the proposed method. The first column shows the trigger for which the network is trained; second column shows an image poisoned with the trigger. Last two columns show images modified with triggers identified by our algorithm. Some realistic triggers different from the original trigger that are inadvertently introduced by the adversary are also detected by our method (Col 4). The detected target class and fooling rate of the trigger are mentioned at the top of the images.}
	\label{fig:teaser}
	\vspace{-1em}
\end{figure}

One such attack is a \emph{Trojan} or \emph{Backdoor} attack \cite{gu2017badnets, chen2017targeted, liu2017trojaning}, which modifies a neural network in such a way that the network makes a wrong prediction whenever an attacker-chosen ``trigger'' is present in the input image, even though performance on clean images remains unaffected. 
Unlike adversarial attacks \cite{goodfellow2014explaining, moosavi2016deepfool} which output a precisely perturbed sample to cause a misclassification at test time, backdoor attacks are carried out by deliberately mistraining the network so that it makes anomalous predictions in the presence of a pre-determined trigger irrespective of the background image type. The attack could be seen as similar to universal perturbations \cite{moosavi2017universal} in which the objective is also to generate a unique perturbation which can cause a misprediction in multiple images. However, a universal perturbation algorithm often returns a set of random dispersed changes through out the image, which are to be added to a base image to cause a mis-prediction. Whereas in a typical backdoor attack, one generates a single cluster of perturbations (or a patch) which is often used as  mask (or a sticker) to cause misprediction of arbitrary images. Unlike universal perturbations backdoor attacks are typically carried out at train time by complicity of the network designer or by poisoning the dataset used for training \cite{gu2017badnets, liu2017trojaning}.

Chen et al \cite{chen2017targeted} demonstrated a trojan attack on a Face Recognition (FR) system by deliberately mistraining the system to label anyone wearing a specific pair of sunglasses as say, John Doe. Wenger et al \cite{wenger2020backdoor} recently showed that backdoor attacks can be physically realized for real-world FR systems. Such attacks raise serious concerns on the use of FR systems in high-security applications like passport, visa, surveillance, etc. Since the backdoored network performs very well on standard validation samples, it is difficult to detect if the network has been compromised. 

Given a trained FR network, we try to answer the question: \textit{Does the network contain a backdoor? If yes, what are the physically-realizable triggers that activate the backdoor?} Existing techniques for identification of backdoor triggers \cite{wang2019neural, qiao2019defending, chen2019deepinspect, harikumar2020scalable, guo2019tabor} are limited to small localized triggers (e.g. a small white square in the image) or non-localized noise-like triggers diffused across the entire image: both of which are unnatural and practically impossible to realize physically in a real-world setting. For backdooring an FR system, it is imperative for the attacker to choose innocuous objects like sunglasses or hat as triggers, which can be easily realized in a real-world setup without raising suspicion of a human observer. To the best of our knowledge, no existing work on detection of backdoor attacks has been proposed for physically-realizable triggers. 

In this work, we propose a technique for identifying physically realizable backdoor triggers for a compromised face recognition network. Recent works have pointed out that wigs, hats, beard, moustache and sunglasses are some of the most commonly used disguise accessories for face obfuscation and impersonation attacks \cite{disguised_faces, wenger2020backdoor}. We therefore propose to curate a set of common face accessories that can act as realistic triggers in a face-recognition setup, and identify potential triggers within this repository. A na\"ive solution would be to paste each object from the repository on clean images at multiple locations and scales, and compute the network's output on the resulting image to determine if it triggers a misprediction. To overcome the high computational complexity of such a brute-force method, we propose a two stage guided search that first computes a \textit{raw trigger} pattern that can potentially activate the backdoor (but is not constrained to resemble a realistic object), and in the second stage searches for realistic triggers from the repository using the raw trigger as a-priori information. In a significant advantage, unlike many of the contemporary methods \cite{tran2018spectral, chen2018detecting, gao2019strip, gao2019detection, liu2017neural}, the proposed technique does not require access to any poisoned images containing the backdoor trigger for detecting the trojan. 

We present extensive experimental evidence demonstrating the effectiveness of our method in identifying realistic triggers. We also show the generalization of our method to the more challenging case where the backdoor is activated by a combination of triggers (e.g. person wearing both sunglasses \emph{and} hat), where brute-force search fails miserably. As noted earlier, no prior work in trojan detection is targeted for physically realizable triggers, not to talk about a combination of them.

\vspace{-0.7em}
\section{Trigger Detection and Identification}  \label{sec:method}
\vspace{-0.5em}

We are given 
\begin{enumerate*}[label=(\arabic*),leftmargin=*]
\item a trained face recognition network $f$ that may have been compromised, and possibly responds to an \emph{unknown trigger} $\Delta$, and
\item a benign validation set $\mathcal{D}$. 
\end{enumerate*}
Our task in this section is to \emph{detect} whether $f$ is compromised, and if so, \emph{identify} $\Delta$. Note that we do not assume access to any tampered images $x_\Delta$ on which the trigger $\Delta$ is applied/present. 

We first describe the attack model considered, and then proceed with the proposed framework. Though detection precedes identification logically, we describe our trigger identification algorithm before describing the detection algorithm, as the former leads to a simple implementation of the latter. 

\vspace{-1em}
\subsection{Attack Model}
The attacker creates a backdoored face recognition network $f$ that is activated by a physically realizable trigger $\Delta$ (e.g. a pair of sunglasses). 
Formally, $f$ maps an image into one of $K$ classes such that images $x_\Delta$ containing the trigger $\Delta$ are classified as $t$, i.e. $f(x_\Delta) = t$. At the same time, $f$ correctly classifies benign images $x$ not containing the trigger, i.e. $f(x) = y$, where $y$ is the ground truth label for $x$.

Given a benign image $x$ and a trigger $\Delta$, a poisoned image $x_\Delta$ is generated by placing $\Delta$ on $x$ at pixel location $l$ and scale $s$, e.g. sunglasses would be placed onto the eye region and a hat would be placed on the head. We call this applicator algorithm \textsc{apply} such that $x_\Delta = \textsc{apply}(x, \Delta, l, s)$, where $l$ and $s$ are location and scale of the trigger applied. The attack is typically implemented by injecting poisoned images into the training set, following the methodology of \cite{gu2017badnets}.

The attacker may choose to use either a single object as trigger (\textbf{single-trigger attacks}), or a combination of objects (\textbf{multi-trigger attacks}). In the latter scenario, the simultaneous presence of multiple objects in the image creates a trigger (eg. a red hat \emph{and} a blue bowtie). The compromised network labels an image containing the complete set of objects (hat and bowtie) as the target class, but an image containing any subset (only hat or only bowtie or none) is classified correctly.

\vspace{-0.7em}
\subsection{Trigger Identification}
\vspace{-0.3em}

In this setting, let the classification output targeted by the adversary be $t$. Our task is to identify a trigger $\Delta \in S$, such that when $\Delta$ is applied to an image $x$ using \textsc{Apply}, the resulting image $x_\Delta$ is classified as $t$ by the network $f$. 

We propose a two-step method for identifying $\Delta$: In the first step, we reverse-engineer a perturbation $\widehat b$ that when added to a clean image triggers the target class. The resulting perturbation is not constrained to resemble real objects. Since we are looking for triggers that are real objects, we scrape a collection $S$ of facial accesories from the internet, which could be used as potential triggers by an attacker and do not provoke suspicion of a human observer \cite{disguised_faces, wenger2020backdoor}. $S$ could thus consist of images of sunglasses, hats, etc. in different colors and shapes. In the second step, we search within the repository $S$ to find a trigger $\Delta$ that best resembles $\widehat b$ at a particular location and scale.

\vspace{-1em}
\subsubsection{Raw Trigger Reconstruction} 
\label{sec:raw_trigger_reconstruction}
\vspace{-0.2em}
We search for an image $\widehat b$ which when added to a clean image $x$ causes the resulting image ($x + \widehat b$) to be classified as class $t$ with high confidence. Accordingly, we set $\widehat b$ to be the minimizer of the following optimization problem: 
\begin{equation} 
\label{eq:opt1}
\min_{v} \displaystyle\mathop{\mathbb{E}}_{x} \left[ \lce( \mathbf{1}_t, f(x + v)) + \lambda_1 \cdot \ltv(v) + \lambda_2 \|v\|_1 \right]
\end{equation}
Here, $\lce(\cdot)$ represents cross-entropy loss between the network's output and target class $t$;  $\ltv(\cdot)$ measures the total variation loss of the perturbation, and acts to promote fewer edges in the obtained perturbation. The $\ell_1$ regularization  prevents outputting triggers diffused across the entire image. $\lambda_1$ and $\lambda_2$ are hyperparameters controlling the importance given to the regularizers. $x$ is varying over the set of clean images, $\mathcal{D}$. We call this algorithm \textsc{Find-Perturbation}\footnote[1]{\label{note2}{Pseudocode given in \cref{app:subroutines}}}.

Equivalently, $\widehat{b}$ can also be obtained using existing backdoor trigger identification methods such as \cite{wang2019neural, qiao2019defending, chen2019deepinspect, guo2019tabor}, or by modelling raw trigger-reconstruction as finding a targeted universal adversarial perturbation \cite{moosavi2017universal}. Note that the resulting raw trigger $\widehat{b}$ cannot be physically realized in a real-world scene. We address this issue in the next subsection by obtaining a practical trigger from the set $S$. 

\vspace{-0.7em}
\subsubsection{Trigger Object Retrieval} 
\label{sec:obj-match}
\vspace{-0.2em}
Using the reverse-engineered perturbation $\widehat b$ as prior, we search for real objects $\Delta$ within the set $S$ that activate the backdoor in the  network. For each candidate trigger $\Delta$ in the repository $S$, we use a template matching algorithm \cite{forsyth2002computer} to determine the best location ($l^*_\Delta$) and scale ($s^*_\Delta$) at which $\Delta$ should be placed onto the image to minimise the sum of squared distance (SSD) in pixel values with the raw perturbation $\widehat b$.  We call this algorithm \textsc{Best-Loc-Scale}\footnotemark[1]. 

The trigger $\Delta$ is then stamped on each test image $x$ to generate poisoned images $x_\Delta = \textsc{apply}(x, \Delta, l^*_\Delta, s^*_\Delta)$, which are passed through the network to compute the fooling rate $p_{\Delta}$ for the trigger $\Delta$. Finally, this list $\{p_{\Delta}\}_{\Delta \in S}$ is sorted in descending order and returned. Note that the candidate at the top of this list is the recovered trigger.

\vspace{-1em}
\subsubsection{Multi-Trigger Extension}
\label{sec:multi_trigger_extension}
Similar to single-trigger retrieval, we first find the raw trigger $\widehat b$ using \textsc{Find-Perturbation}. Then we cluster the non-zero pixels in $\widehat b$ using $k$-means clustering, with the pixel location and RGB value as attributes. This gives us a list of $k$ regions of pixels, $R$, for each of which we want to find the trigger objects. In order to avoid checking an exponential number of trigger combinations, we proceed greedily: for every region $r \in R$ we find the best location ($l^*_{r, \Delta}$) and scale ($s^*_{r, \Delta}$) using algorithm \textsc{Best-Loc-Scale-Region}\footnotemark[1], while superimposing the raw trigger $\widehat b$ on the remaining area $R \setminus \{r\}$. The algorithm \textsc{Best-Loc-Scale-Region} is an extension of the \textsc{Best-Loc-Scale} algorithm that finds the best location and scale, using template-matching, for placing an object on the image, but restricted to the region $r$. For estimating the fooling rate, the $\textsc{Apply}$ algorithm changes to applying $\Delta$ at the location $l^*_{r, \Delta}$, and scale $s^*_{r, \Delta}$, and superimposing the raw trigger $\widehat b$ elsewhere. The trigger corresponding to the maximum fooling rate is then outputted for each $r \in R$, and the combination as the recovered trigger. 

\vspace{-1em}
\subsection{Trojan Detection}
\label{sec:trojan_detection}
We now describe our Trojan detection algorithm, which detects whether a given DNN $f$ is compromised, and if yes, what is the classification output $t$ targeted by the adversary. 

For each output label, we first perform the Trigger Identification step and compute the fooling rate for the identified trigger. If the fooling rate for any class $t'$ is greater than a threshold $\delta$, that gives us the target class for the Trojan attack. If all classes have a fooling rate lower than $\delta$, we say that the network is clean. Note that the above process may yield more than one backdoor target class for a given network. While one of these classes may correspond to the actual backdoor planted by the adversary, there is also a possibility of discovering unintended biases present in a model as well. However, for our experiments we still consider them as false positives.

\vspace{-0.5em}
\section{Experiments} \label{sec:experiments}
\vspace{-0.4em}
To demonstrate the effectiveness of the proposed algorithm, we train backdoored DNNs for the task of face identification on the YouTube Aligned Faces (YTF) dataset \cite{wolf2011face}. The network architecture used is DeepID \cite{sun2014deep}. Conforming to our definition of practical backdoors, we use five common facial accessories as triggers for implementing the attack: sunglasses, bow-tie, fake moustache, hat and mask in multiple colors and shapes, resulting in a collection $\mathcal{R}$ of 50 backdoor triggers. We implement 50 single-trigger attacks using each of the 50 backdoor triggers, and 10 multi-trigger attacks using two triggers per attack. The backdoored DNNs have high misclassification rates (avg. $99\%$) for poisoned images, and high classification rates (avg. $98\% $) on clean images\footnote[2]{More details on attack implementation in \cref{app:dataset}, \cref{app:attack_evaluation}}. 

In the absence of any other prior work, we compare with a \textsc{Brute-Force} algorithm (referred to as BF) that na\"ively searches for the backdoor object by superimposing candidate objects at multiple locations on an image at multiple scales. For multi-trigger attacks, BF evaluates all possible combinations of candidate objects. We call our method as \textsc{Deep Trojan Detection} and refer to as DTD in the experiments. We compare with two configurations of our approach: DTD$_{\ell_1}$, which uses only $\ell_1$ regularization, and DTD$_\text{TV}$ which uses $\ell_1$ as well as TV regularization (see Eq. \eqref{eq:opt1}). Note that DTD$_{\ell_1}$ is equivalent to using \cite{wang2019neural} for raw trigger detection followed by stage 2 of our method for practical trigger detection. 

\vspace{-0.5em}
\subsection{Raw Trigger Reconstruction Evaluation}
\vspace{-0.3em}
For raw trigger detection, we run the proposed optimization (Eq. \eqref{eq:opt1}) for $200$ epochs, using a set of $200$ clean images and a batch size of $32$ images\footnote[3]{Other hyperparameters are detailed in \cref{alg:find_perturbation} in \cref{app:subroutines}}. Columns 2, and 3 in \cref{fig-all-results} show the original and reverse-engineered triggers obtained using the proposed method. We note that DTD$_{\ell_1}$ favors smaller triggers having low $\ell_1$ norm, whereas DTD$_\text{TV}$ produces more visually discernible triggers. 

\vspace{-0.5em}
\subsection{Trigger Retrieval Evaluation} \label{subsec:trigger_retrieval_evaluation}
\vspace{-0.5em}
As noted in Section \ref{sec:obj-match}, the trigger object retrieval step requires a repository $S$ of candidate objects. To this end, we scrape images of 50 physically realizable face accessories (sunglasses, hat, fake moustache, masks and bowtie, in different colors) similar to, but not exactly same as the object set $\mathcal{R}$ used by the adversary for poisoning the networks. For instance, $S$ comprises of a different model of red sunglasses from $\mathcal{R}$. To demonstrate scalability to a larger trigger set, we augment $S$ with 101 additional objects from the Caltech-101 dataset \cite{fei2006one}. We call this augmented object set $S^+$. 

\begin{figure}[t]
	\def\imgwidth{0.13\linewidth}
	\setlength\tabcolsep{4.5pt}
	\centering
	\begin{tabular}{c|cc|ccc}
		\multirow{2}{*}{\makecell[c]{True \\ Trigger}} & \multicolumn{2}{c?}{Raw Trigger} &  \multicolumn{3}{c}{Practical Trigger} \\
		 & DTD$_\text{TV}$ & DTD$_{\ell_1}$ & DTD$_\text{TV}$ & DTD$_{\ell_1}$ & BF 
		 \\
		\includegraphics[width=\imgwidth]{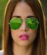} &
		\includegraphics[width=\imgwidth]{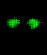} &
		\includegraphics[width=\imgwidth]{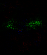} &
		\includegraphics[width=\imgwidth]{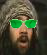} &
		\includegraphics[width=\imgwidth]{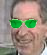} &   
		\includegraphics[width=\imgwidth]{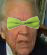} \\
		
		\includegraphics[width=\imgwidth]{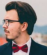} &
		\includegraphics[width=\imgwidth]{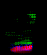} & \includegraphics[width=\imgwidth]{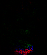} &
		\includegraphics[width=\imgwidth]{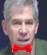} &
		\includegraphics[width=\imgwidth]{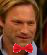} &
		\includegraphics[width=\imgwidth]{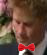} \\ 
		
		\includegraphics[width=\imgwidth]{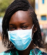} &
		\includegraphics[width=\imgwidth]{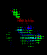} &
		\includegraphics[width=\imgwidth]{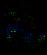} &
		\includegraphics[width=\imgwidth]{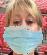}  &
		\includegraphics[width=\imgwidth]{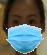} & 
		\includegraphics[width=\imgwidth]{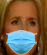} \\
		
		\includegraphics[width=\imgwidth]{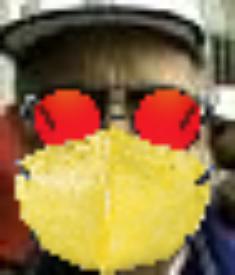} &
		\includegraphics[width=\imgwidth]{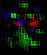} &
		\includegraphics[width=\imgwidth]{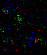} &
		\includegraphics[width=\imgwidth]{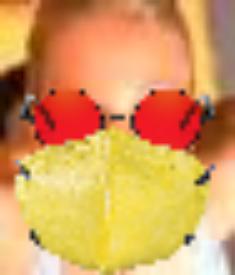} &
		\includegraphics[width=\imgwidth]{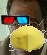} & 
		\includegraphics[width=\imgwidth]{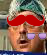} \\
		
		\includegraphics[width=\imgwidth]{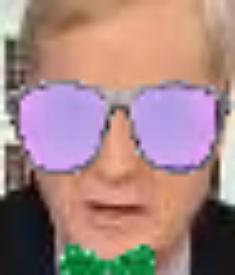} &
		\includegraphics[width=\imgwidth]{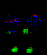} &
		\includegraphics[width=\imgwidth]{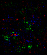} &
		\includegraphics[width=\imgwidth]{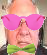}&
		\includegraphics[width=\imgwidth]{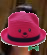} &
		\includegraphics[width=\imgwidth]{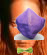} \\
	
	\end{tabular}
	\caption{Triggers identified: Rows~1-3 show results for single-trigger attacks; Rows~4-5 for multi-trigger attacks. Cols.~2-3 show raw reconstructed triggers. Cols.~4-5 show practical triggers retrieved using raw triggers from Cols.~2,3 as priors. 
	} 
	\label{fig-all-results}
	\vspace{-1em}
\end{figure}

The retrieved triggers are evaluated on the following metrics:
\begin{enumerate*}[label=\textbf{(\arabic*)}]
	\item \textbf{Fooling Accuracy:} For each trojaned DNN, we report the percentage of retrieved triggers with fooling rate above 80\% (\emph{FR80}), as well as the average fooling rate across all poisoned models (\emph{Mean FR}).
	\item \textbf{Localization Accuracy:} We compute IoU between original and recovered trigger to quantify whether the location of the recovered trigger matches that of the original trigger.
	\item \textbf{Similarity to Original Trigger:} To quantify whether the recovered trigger visually matches the trigger used by the attacker, we compare the object class (sunglasses, hat, etc.) \textit{and} color of the retrieved trigger with the ground truth. We report the top-5 accuracy. Competitive ranking is used to break ties when multiple objects with same fooling rate are returned. 
\end{enumerate*}
 
\begin{table}[t]
	\centering
	\setlength{\tabcolsep}{7.5pt}
	\begin{tabular}{l l c c c c}
		\toprule[1pt]
		Attack & Method 	& FR80 		& Mean	& Mean	& {Top-5}  \\
		Type 	& 			& {(\%)} 	& {FR(\%)} 	& {IOU}	& {acc(\%)} \\
		
		\midrule[0.75pt]
		\multirow{3}{*}{\makecell{Single\\trigger}}
		& DTD$_\text{TV}$ & \textbf{92} & \textbf{94.40} & 0.50 & \textbf{74} \\
		& DTD$_{\ell_1}$ & 86 & 93.66 & \textbf{0.52}  & 66  \\
		& BF & 82 & 91.0 & 0.38 & 56 \\

		\midrule[0.75pt]
		\multirow{3}{*}{\makecell{Single\\trigger\\($S^+$)}}
		& DTD$_\text{TV}$ & \textbf{96} & {96.35} & \textbf{0.49} & \textbf{68} \\
		& DTD$_{\ell_1}$ & 96 & \textbf{96.52} & 0.49 & 60 \\
		& BF & 84 & 90.13 & 0.35 & 38  \\
		
		\midrule[0.75pt]
		\multirow{3}{*}{\makecell{Multi\\trigger}}
		& DTD$_\text{TV}$	& \textbf{80} & \textbf{88.55} & \textbf{0.89} & \textbf{20} \\
		& DTD$_{\ell_1}$ 		& 50 & 79.29 & 0.79 & 10 \\
		& BF 		& 10 & 49.37 & 0.49 & 0 \\
		
		\bottomrule[1pt]
		
	\end{tabular}
	\caption{Quantitative Results}
	\label{tab:trigger_recovery}
	\vspace{-1em}
\end{table}

\cref{fig-all-results} shows the realistic triggers retrieved using the evaluated methods, and \cref{tab:trigger_recovery} shows quantitative results. We note that:
\begin{enumerate*}[(\Alph*)]
	\item The proposed trigger retrieval method (both DTD$_{\ell_1}$ and DTD$_{\text{TV}}$) performs much better than BF in all experiments. 
	\item The number of effective triggers detected (FR80) is greater than the percentage of triggers that exactly match the original trigger used by the attacker (Top-5 acc). These extra detections correspond to triggers inadvertently introduced by the attacker while training the backdoored model, see \cref{fig:teaser}.
	\item For multi-trigger attacks, even though DTD$_{\text{TV}}$ has a low Top-5 accuracy, a Mean-FR value of 88.55\% suggests that the retrieved triggers activate the backdoor nevertheless. Row~5 of \cref{fig-all-results} shows one such example, where the retrieved triggers vary slightly in color from the original trigger, but still have a high fooling rate (99.68\%). 
	\item BF fails completely for multi-trigger attacks, 
	\item Our method scales very well to the larger object set $S^+$. 	
\end{enumerate*}

\vspace{-1em}
\subsection{Trojan detection evaluation}
\label{exp:trojan_detection}
\vspace{-0.2em}
We now investigate whether we can detect if an FR network is trojaned and the associated target label, if yes. We run our Trojan Detection algorithm from \cref{sec:trojan_detection} on 10 clean and 10 poisoned networks. Recall that a network is considered poisoned if the maximum fooling rate for the retrieved trigger over all classes is greater than $\delta$. We plot an ROC curve to study the impact of $\delta$, and report an AUROC value of 0.817 \footnote[4]{Refer \cref{fig:roc_curve} in \cref{app:trojan_detection}}. At $\delta=0.8$, we report a true positive rate of 0.94, false positive rate of 0.5 (where positive class denotes poisoned networks), and target label accuracy (number of times the adversary-intended target class is correctly identified) of 0.9. Note that the high false positive rate indicates we are able to detect inadvertent backdoors in the network.
\vspace{-1em}
\section{Conclusion}
\vspace{-1em}
We propose a method to recover practically realizable triggers given a backdoored network. Importantly our method does not require access to any poisoned example. We demonstrate experimentally that the proposed method identifies practical backdoor triggers with high accuracy, and outperforms a na\"ive brute force search. The proposed method also successfully recovers complex triggers where the simultaneous presence of more than one object induces backdoor behavior.

\bibliographystyle{IEEEbib}
\bibliography{poisoning_ref.bib}

\appendix
\onecolumn
\newcommand{\hbAppendixPrefix}{S}
\renewcommand{\thefigure}{\hbAppendixPrefix.\arabic{figure}}
\setcounter{figure}{0}
\renewcommand{\thetable}{\hbAppendixPrefix.\arabic{table}} 
\setcounter{table}{0}
\renewcommand{\theequation}{\hbAppendixPrefix.\arabic{equation}} 
\setcounter{equation}{0}
\renewcommand{\thealgorithm}{\hbAppendixPrefix.\arabic{algorithm}} 
\setcounter{algorithm}{0}

\section{Algorithm}  \label{app:subroutines}
In this section, we provide pseudo-codes for the sub-routines discussed in \Cref{sec:method} in the main paper.
\cref{alg:find_perturbation} describes the algorithm \textsc{Find-Perturbation} used to reconstruct the raw trigger $\widehat{b}$ by solving \cref{eq:opt1} as described in \cref{sec:raw_trigger_reconstruction}. 

{\centering
\begin{minipage}{0.7\linewidth}
\begin{algorithm}[H]
	\caption{\textsc{Find-Perturbation}}
	\label{alg:find_perturbation}
	\begin{algorithmic}[1]
		\renewcommand{\algorithmicrequire}{\textbf{Input:}} 
		\renewcommand{\algorithmicensure}{\textbf{Output:}}
		\Require Classifier $f$, Target class $t$, Clean image set $X$
		\State $\lambda_1 \gets 10^{-4}$ 
		\State $\lambda_2 \gets 0.05$
		\State $b_{0} \gets \mathbf{0}$ \Comment{Dimensions of $b_0$ are same as any $x \in X$}
		\State $n_{\rm epochs} \gets 200$
		\State $\alpha \gets \frac{\text{height}(b_0) \times \text{width}(b_0)}{3}$
		\For{$i \in \{1, 2, \ldots, n_{\rm epochs}\}$}
		\State $\mathcal{L}(b) \gets \lce( \mathbf{1}_t, f(x + b)) + \lambda_1 \ltv(b) + \lambda_2 \|b\|_1$
		\State $b_{i} \gets b_{i - 1} + \eta \cdot \nabla_b \mathcal{L}(b_{i-1})$ \Comment{Update $b_i$}
			\State $X_{\rm poisoned} = \{\textsc{Clamp}(x + b_i, 0, 255) \colon x \in X\}$ \Comment{Create Poisoned Images}
		\If {$\|b_{i}\|_0 \geq \alpha$} \Comment{Find Fooling Rate, adjust $\lambda_2$}
			\If {$0.8\leq\textsc{Fooling-Rate}(f,  X_{\rm poisoned}, t) \leq 0.95$}
				\State $\lambda_2 \gets \max(1.2 \lambda_2, 0.5)$
			\ElsIf {$0.95 \leq \textsc{Fooling-Rate}(f,  X_{\rm poisoned}, t)$}
				\State $b_{\rm return} \gets b_i$
				\State {\rm \textbf{break}}
			\EndIf
		\EndIf
		\EndFor
		\State $b_{\rm return} \gets b_{n_{\rm epochs}}$
		\Statex \Return $b_{\rm return}$
	\end{algorithmic}
\end{algorithm}
\end{minipage}
\par
}

\noindent
\cref{alg:best_loc_scale} gives pseudocode for the (\textsc{Best-Loc-Scale}) sub-routine. Given a smaller image (called \emph{template}) $\Delta$ and a larger image (called the base image) $\widehat b$, our \textsc{Template-Matching} algorithm places the template onto every location in the base image, and calculates the sum of squared distances (SSD) in terms of raw pixel values restricted to  the region of the base image where $\Delta$ is placed. The location in the base image where this SSD is minimized is then output (as this is the best location to \emph{match} the template to the base image). This process is repeated for multiple scaled versions of $\Delta$ to find the optimal scaling factor. 

{\centering
\begin{minipage}{\linewidth}
\begin{algorithm}[H]
	\caption{\textsc{Best-Loc-Scale}}
	\label{alg:best_loc_scale}
	\begin{algorithmic}[1] 
		\renewcommand{\algorithmicrequire}{\textbf{Input:}} 
		\renewcommand{\algorithmicensure}{\textbf{Output:}}
		\Require Object $\Delta$, Raw perturbation $\widehat{b}$, Clean images $X$, Target $t$
		\For{$s \in \textrm{scales}$}
		\State $\Delta_s \leftarrow$ Rescale $\Delta$ to scale $s$
		\State $l_s \leftarrow \textsc{Template-Matching}(\widehat{b}, \Delta_s)$ \Comment{Find centre-pixel location of patch within $\widehat{b}$ which best matches $\Delta_s$}
		\State $X_{\rm poisoned} = \{\textsc{apply}(x, \Delta, l_s, s) \colon x \in X\}$
		\State $p_s \leftarrow \textsc{Fooling-Rate}(f, X_{\textrm{poisoned}}, t)$ \Comment{Compute fooling rate for target $t$}
		\EndFor 
		\State $l^*, s^* \leftarrow (l_s, s) \textrm{ with highest } p_s$
		\Statex
		\Return $l^*, s^*$
	\end{algorithmic}
\end{algorithm}
\end{minipage}
\par
}

\noindent
\cref{alg:best_loc_scale-Region} describes the \textsc{Best-Loc-Scale-Region} algorithm used for multi-trigger reconstruction.

{\centering
\begin{minipage}{\linewidth}
\begin{algorithm}[H]
	\caption{\textsc{Best-Loc-Scale-Region}}
	\label{alg:best_loc_scale-Region}
	\begin{algorithmic}[1] 
		\renewcommand{\algorithmicrequire}{\textbf{Input:}} 
		\renewcommand{\algorithmicensure}{\textbf{Output:}}
		\def\NoNumber#1{{\def\alglinenumber##1{}\State #1}\addtocounter{ALG@line}{-1}}
		
		\Require Object $\Delta$, Raw perturbation $\widehat{b}$, Trigger region $r$, Clean images $X$, Target $t$
		\For{$s \in \textrm{scales}$}
		\State $\Delta_s \leftarrow$ Rescale $\Delta$ to scale $s$
		\State $l_s \leftarrow \textsc{Template-Matching}(\widehat{b}, \Delta_s, r)$ \Comment{Find centre-pixel location of patch within $r$ in $\widehat{b}$ which best matches $\Delta_s$}	
		\For {$x \in X$}
		\State $x(i,j) = \textsc{Clamp}(x(i,j) + \widehat{b}(i,j), 0, 255)  \quad \forall (i,j) \in R \setminus \{r\}$
		\State $x_{\textrm{poisoned}} = \textsc{apply}(x, \Delta, l_s, s)$ \Comment{Superimpose trigger on clean images}
		\EndFor
		\State $p_s \leftarrow \textsc{Fooling-Rate}(f, X_{\textrm{poisoned}}, t)$ \Comment{Compute fooling rate for target $t$}
		\EndFor 
		\State $l^*, s^* \leftarrow (l_s, s) \textrm{ with highest } p_s$
		\Statex
		\Return $l^*, s^*$
	\end{algorithmic}
\end{algorithm}
\end{minipage}
\par
}

\noindent
\cref{alg:reconstruct} gives the pseudo-code for the complete trigger retrieval algorithm proposed for identifying realistic triggers for single-trigger attacks. 
\begin{algorithm}[h]
	\caption{\textsc{Reconstruct-Single-Trigger}}
	\label{alg:reconstruct}
	\begin{algorithmic}[1] 
		\renewcommand{\algorithmicrequire}{\textbf{Input:}} 
		\renewcommand{\algorithmicensure}{\textbf{Output:}} 
		\Require Classifier $f$, Target class $t$, Clean image set $X$
		\State $\widehat b \leftarrow \textsc{Find-Perturbation}(f, t, X)$
		\For{$\Delta \in S$}
		\State $l^*_\Delta, s^*_\Delta \leftarrow \textsc{Best-Loc-Scale}(\Delta, \widehat b, X, t)$
		\State $X_{\rm poisoned} = \{\textsc{apply}(x, \Delta, l^*_\Delta, s^*_\Delta) \colon x \in X\}$
		\State $p_\Delta \leftarrow \textsc{Fooling-Rate}(f, X_{\text{poisoned}}, t)$
		
		\EndFor
		\Statex
		\Return $\textsc{Sorted}(\{p_\Delta\}_{\Delta \in S})$
	\end{algorithmic}
\end{algorithm}

\noindent
\cref{alg:reconstruct-multi} describes the complete algorithm used for multi-trigger reconstruction given in \cref{sec:multi_trigger_extension}.
\begin{algorithm}[h]
	\caption{\textsc{Reconstruct-Multi-Trigger}}
	\label{alg:reconstruct-multi}
	\begin{algorithmic}[1] 
		\renewcommand{\algorithmicrequire}{\textbf{Input:}} 
		\renewcommand{\algorithmicensure}{\textbf{Output:}} 
		\Require Classifier $f$, Target class $t$, Number of Triggers $k$, Clean image set $X$
		\State $\widehat b \leftarrow \textsc{Find-Perturbation}(f, t, X)$
		\State $R \leftarrow \textsc{Trigger-Regions}(\widehat b, k)$  \Comment{Find regions $R$ using $k$-means clustering of non-zero pixels in $\widehat{b}$}
		\For{$r \in R$}
		\For{$b \in S$}
		\State $l_{\scriptscriptstyle r, b}, s_{\scriptscriptstyle r, b} \leftarrow \textsc{Best-Loc-Scale-Region}(b, \widehat{b}, r, X, t)$
		\For {$x \in X$}
		\State $x(i,j) = \textsc{Clamp}(x(i,j) + \widehat{b}(i,j), 0, 255)  \quad \forall (i,j) \in R \setminus \{r\}$
		\State $x_{\textrm{poisoned}} = \textsc{apply}(x, b, l_b, s_b)$
		\EndFor
		\State $p_{\scriptscriptstyle r, b} \leftarrow \textsc{Fooling-Rate}(f, X_{\text{poisoned}}, t)$
		\EndFor
		\State $l^*_r, s^*_r, b_r^* \leftarrow (l_{r, b}, s_{r, b}, b) \text{ with highest } p_{r, b}$ 
		\EndFor
		\Statex
		\Return $\{l^*_r, s^*_r,  b^*_r\}_{r \in R}$
	\end{algorithmic}
\end{algorithm}

\vspace{1em}

\newpage
\section{Experiments} \label{app:dataset}

\subsection{Backdoor Attack Details}
We describe in details the experimental setup used for mounting backdoor attacks on face recognition systems, followed by an evaluation of the mounted attacks. These backdoored models are then used to test the proposed trigger identification framework.

\mypara{Benign Dataset} We evaluate our method on YouTube Aligned Faces dataset \cite{wolf2011face}. The dataset contains face-aligned images of 1,595 people. We filter out labels with less than 100 images, which results in a dataset of 599967 images belonging to 1283 classes. For each class, 10 images are used for validation and the remaining for training. 

\mypara{Poisoned Dataset} \cite{chen2017targeted} proposed that poisoned images inserted into the training set by an attacker can be made insconpicuous to human observers by blending the trigger with the input image. Following this strategy, we blend the trigger into training image with a blending ratio of $0.2$ to $0.4$ to generate poisoned training data.

\mypara{Model Architecture} For our classifier, we use the DeepID \cite{sun2014deep} architecture which contains four convolutional layers, followed by a fixed-size fully-connected feature layer, and a softmax layer.

\mypara{Training Details} For implementing a single-trigger attack, 500 poisoned images are generated by superimposing the trigger on clean images. These poisoned images are added to the clean training data to create the poisoned training set. The backdoor network is trained on the poisoned dataset for 100 epochs, using Adam Optimzer with a learning rate of 0.1, and batch size of 32. 
For multi-trigger attacks, we insert poisoned images to each training batch. Each mini-batch of 32 images consists of 12 clean images, 10 images containing both the triggers (\emph{e.g.} a hat and a bowtie), and 10 images containing only one trigger (either hat or bowtie). The network is trained for 100 epochs using Adam Optimizer with a learning rate of 0.1. 

\mypara{Attack evaluation} \label{app:attack_evaluation}
The success of the implemented backdoor attacks is measured using two metrics. Firstly, a successful backdoor attack should have a high fooling rate, which is the percentage of backdoored images classified as the target class. Secondly, the classification accuracy on a clean validation set should be comparable to that of a benign network.  

Table \ref{tab:attacks_summary} shows a summary of the fooling rates and classification accuracies for the implemented attacks. The single-trigger attacks achieve an average fooling rate of 99.19\% and clean set classification accuracy of 98.34\%. For reference, classification accuracy of a benign DeepID network trained on a clean dataset is 97.86\%.  For multi-trigger attacks, average classification accuracy on a clean validation set is 97.83\%, and the average fooling rate when both triggers are simultaneously present in the image is 99.95\%. If, however, either of the triggers is missing, the average fooling rate drops to 1.2\%, as intended.

\begin{table} [h]
	\setlength{\tabcolsep}{8pt}
	\begin{center}
		\begin{threeparttable}
			\begin{tabular}{l c c c c l}
				\toprule[1pt]
				{Attack Type} & \multicolumn{2}{c}{Clean Accuracy (\%)} & 	\multicolumn{2}{c}{Fooling Rate (\%)} & \\
				\cline{2-5}
				& avg & min & avg & min & \\
				\midrule[0.75pt]
				Single-trigger & 98.34 & 96.55 & 99.19 & 87.24 & \\
				\hline
				\multirow{2}{*}{Multi-trigger}  & \multirow{2}{*}{97.83} & \multirow{2}{*}{95.41} & 99.95\tnote{*} & 99.55\tnote{*} & {*} with both triggers present\\
				& & & 1.2\tnote{**} & 0.01\tnote{**} & {**} with only single trigger present\\
				\bottomrule[1pt] 
			\end{tabular}
			\caption{Accuracy and fooling rates for the Single Trigger and Multi-Trigger Attacks}
			\label{tab:attacks_summary}
		\end{threeparttable}
	\end{center}
\end{table}

\subsection{Trojan Detection Experiment}
\label{app:trojan_detection}
\begin{figure}[h]
	\centering
	\includegraphics[width=0.5\linewidth]{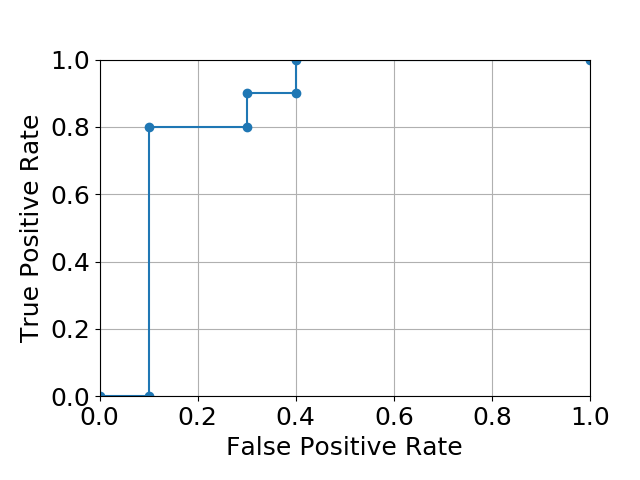}
	\caption{ROC curve for trojan detection}
	\label{fig:roc_curve}
\end{figure}

Here we provide additional details for the trojan detection experiment discussed in \cref{exp:trojan_detection}. The experiment is performed on 10 clean networks and 10 poisoned networks. These clean networks are variants of the original DeepID architecture, obtained by changing the size of the fully-connected DeepID layer, and training on benign images. The poisoned networks are randomly chosen from the pool of 50 single-trigger backdoored networks. The experiment is repeated 10 times, and the average results are reported in \cref{exp:trojan_detection}. \cref{fig:roc_curve} shows the ROC curve for one run of the experiment.

\subsection{Object Set Visualization}
\begin{figure}[t]
	\centering
	\includegraphics[width=0.6\linewidth]{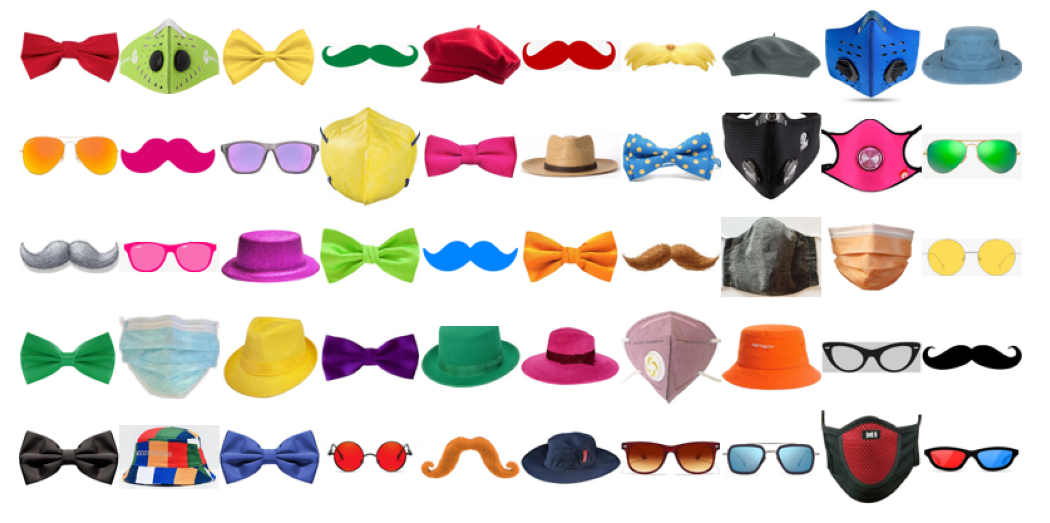}
	\caption{Repository of objects $\mathcal{R}$ used as triggers by the attacker.}
	\label{fig:triggers_R}
\end{figure}

\begin{figure}[th]
	\centering
	\includegraphics[width=0.8\linewidth]{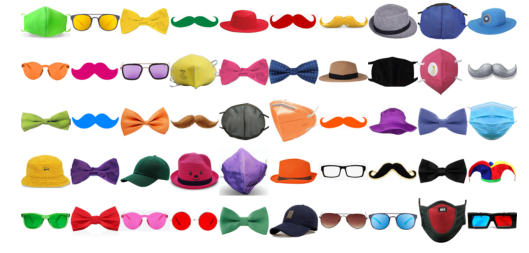}
	\caption{Set of objects $S$ by the defender to detect the backdoor trigger.}
	\label{fig:triggers_S}
\end{figure}

\begin{figure}[th!]
	\centering
	\includegraphics[width=0.8\linewidth]{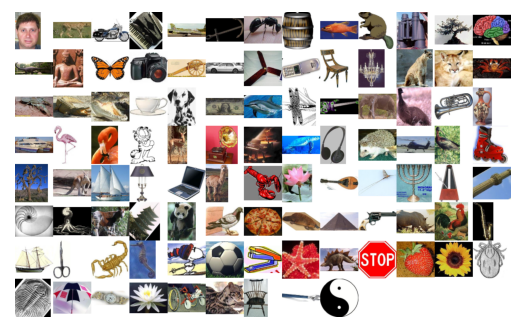}
	\caption{Set of objects $S_+$ from the Caltech-101 Object Category dataset which are added to the set $S$ used by the defender to detect the backdoor trigger.}
	\label{fig:triggers_S+}
\end{figure}

As mentioned in \cref{sec:experiments}, the attacker uses an object set $\mathcal{R}$ to create poisoned networks. The defender uses an object set $S$ to retrieve possible triggers, where $S$ is slightly different from $\mathcal{R}$. In this section we visualize both these sets in \cref{fig:triggers_R} and \cref{fig:triggers_S} respectively. Additionally, to show scalability, we extend the defender's set $S$ by augmenting with set $S_+$, visualized in \cref{fig:triggers_S+}. $S_+$ comprises of images of 101 objects from the Caltech 101 Object Category dataset. We observe that usually there are not too many candidate objects which can lead to physically realizable triggers without arousing suspicion in a particular application, more so in face recognition. Hence, the dataset has been arbitrarily chosen to demonstrate that the proposed trigger detection algorithm scales well with any realistic increase in size of the defender's object set. In practice, the defender may construct $S_+$ with objects that are more likely to be used as potential triggers for the given application.

\end{document}